\documentclass[11pt,a4paper]{article}
\usepackage[hyperref]{emnlp2020}
\usepackage{times}
\usepackage{latexsym}

\usepackage{graphicx}
\usepackage{multirow}
\usepackage{amsmath}
\usepackage{tabularx}
\usepackage{url}
\usepackage{float}
\usepackage{tikz-qtree}
\usepackage{subfig}
\usepackage{pgfplots}
\pgfplotsset{compat=1.16}
\usepackage{latexsym}

\usepackage{microtype}

\aclfinalcopy 

\title{Large Discourse Treebanks from Scalable Distant Supervision}

\author{Patrick Huber and Giuseppe Carenini\\
  Department of Computer Science \\
  University of British Columbia \\
  Vancouver, BC, Canada, V6T 1Z4 \\
  {\tt \{huberpat, carenini\}@cs.ubc.ca}}

\date{}

\begin{document}
\maketitle

\section{Introduction}
Discourse parsing is an important upstream task in Natural Language Processing (NLP) with strong implications for many real-world applications, for example sentiment analysis \cite{bhatia2015better,nejat2017exploring,hogenboom2015using}, text classification \cite{ji2017neural} and summarization \cite{gerani2014abstractive}, just to name a few. Despite this widely recognized role of discourse parsing in NLP, most recent discourse parsers (and consequently downstream tasks) still rely on small scale human annotated discourse treebanks (such as RST-DT \cite{carlsonbuilding}, Instr-DT \cite{subba2009effective} or PDTB \cite{prasadpenn}), trying to infer general-purpose discourse structures from very limited data in a few narrow domains. While in principle diverse and large discourse treebanks could be annotated, the process is expensive, tedious and does not scale.

To overcome this dire situation and allow discourse parsers to be trained on larger, more diverse and domain independent datasets, we propose a framework to generate ``silver-standard" discourse trees from distant supervision on the auxiliary task of sentiment analysis\footnote{This work has been partly published and is partly under submission.}. Our approach explicitly models discourse structures, distinguishing it from previous work on implicit discourse modelling, such as \citet{liu2019single}, previously criticized in \citet{ferracane2019evaluating}.

\section{Distant Supervision Approach}
\begin{figure}[H]
  \centering
  \includegraphics[width=0.50\linewidth]{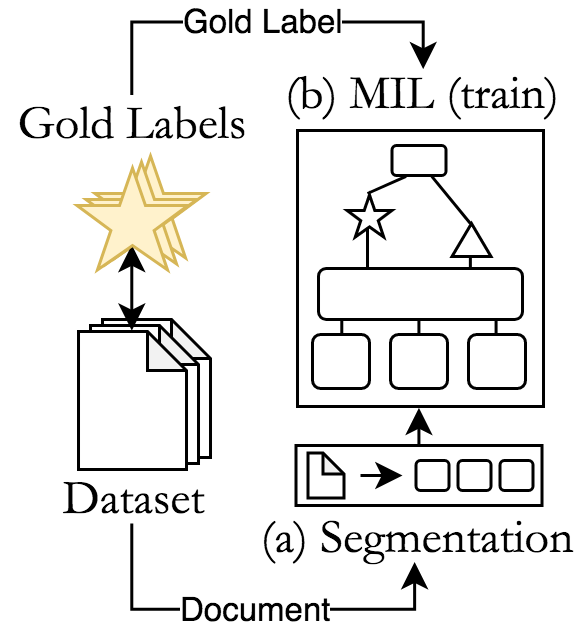}
\caption{First stage, training the MIL model on the Yelp'13 \cite{tang2015document} document-level sentiment prediction task.}
\label{fig:overall_1}
\end{figure}

\begin{figure}[H]
  \centering
  \includegraphics[width=.9\linewidth]{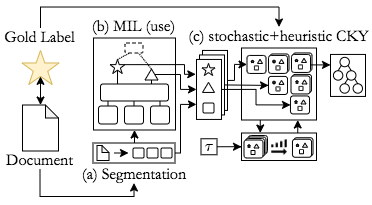}
\caption{Second stage, using the neural MIL model to retrieve fine-grained sentiment and attention scores (star/triangle), used for the stochastically inspired heuristic CKY computation, efficiently generating large discourse trees using an exploration/exploitation trade-off.}
\label{fig:overall_2}
\end{figure}

The overview of our approach, separated in two distinct phases, is shown in Figures \ref{fig:overall_1} and \ref{fig:overall_2}. In the first phase (Figure \ref{fig:overall_1}), we train the neural Multiple-Instance Learning (MIL) model by \citet{angelidis2018multiple} to infer EDU-level sentiment and

attention scores from the document-level gold-label sentiment of the Yelp'13 corpus \cite{tang2015document}. Subsequently (see Fig. \ref{fig:overall_2}), we use the fine-grained EDU-level information to run a CKY-style dynamic programming model, inferring the discourse structure of the document by optimizing the distance between the document-level gold-label sentiment and the discourse-aggregated sentiment prediction.
We call the discourse treebank generated by this approach \textit{Yelp13-DT}.

\begin{table*}
\centering
\scalebox{1}{
\begin{tabular}{|l|r r|r r|}
\hline
\multirow{2}{*}{Approach} & \multicolumn{2}{c|}{Structure} & \multicolumn{2}{c|}{Nuclearity}\\
 & RST-DT & Instr-DT & RST-DT & Instr-DT\\
\hline \hline 
\multicolumn{5}{|c|}{\textbf{Intra-Domain} Evaluation}\\
\hline
CODRA\shortcite{joty2015codra} & 83.84 & \textbf{82.88}  & 68.90 & \textbf{64.13}\\
Two-Stage\shortcite{wang2017two} & \textbf{86.00} & 77.28 & \textbf{72.40} & 60.01\\
\hline \hline
\multicolumn{5}{|c|}{\textbf{Inter-Domain} Evaluation}\\
\hline
Two-Stage\textsubscript{RST-DT} & $\times$ & 73.57 & $\times$ & 49.78\\
Two-Stage\textsubscript{Instr-DT} & 74.32 & $\times$ & 44.68 & $\times$\\
Two-Stage\textsubscript{Yelp13-DT} & 76.41 & 74.14 & -- & --\\
Two-Stage\textsubscript{MEGA-DT} & \textsuperscript{$\dagger$}\textbf{77.82} & \textsuperscript{$\dagger$}\textbf{75.18} & \textbf{44.88} & \textsuperscript{$\dagger$}\textbf{54.87}\\
\hline
\end{tabular}}
\caption{Results of the average micro precision measure for structure- and nuclearity-prediction,
evaluated on the RST-DT and Instr-DT corpora. Inter-domain subscripts identify the training set. Inter-domain results averaged over 10 independent runs. Best performance per sub-table is \textbf{bold}. (\textsuperscript{$\dagger$} statistically significant with p-value $\leq .05$ to the best inter-domain baseline (Bonferroni adjusted), $\times$ not feasible combinations, -- Not pursued)}
\label{tab:final}
\end{table*}

While the CKY approach limits the time-complexity of the computation to $O(n^3)$, the space-complexity grows according to the Catalan number, making the prediction of discourse structures for long documents intractable. We overcome this limitation with a heuristic beam search approach, empirically setting the beam-size to 10 trees per CKY cell. For subtrees on low levels of the discourse tree the overall document sentiment is not necessarily a good pruning criteria. We therefore allow deep subtrees to be more diverse by adding an exploration/exploitation trade-off, as commonly used in Reinforcement Learning (RL). With these two extensions of the standard CKY approach, we can explore additional discourse attributes during the CKY process, such as the ternary nuclearity attribute. Furthermore, nearly arbitrary long documents can be effectively and efficiently processed, significantly expanding the applicability of the approach to new domains. We call a second discourse treebank, generated with the additional heuristic and stochastic components and extended with nuclearity computations \textit{MEGA-DT}.

\section{Results}
We evaluate our ``silver-standard" discourse treebanks on the inter-domain discourse parsing task. We therefore train the top-performing \textit{TwoStage} parser \cite{wang2017two} on the \textit{Yelp13-DT} and \textit{MEGA-DT} discourse-annotated review corpora and evaluate the performance on the commonly used \textit{RST-DT} and \textit{Instr-DT} treebanks in the news- and instructions-domain (see bottom rows in the inter-domain evaluation sub-table of Table \ref{tab:final}). We compare the performance against models trained and evaluated on different domain gold-standard treebanks (top rows in the inter-domain evaluation sub-table). We show common intra-domain discourse parsing results in the first sub-table of Table \ref{tab:final} for comparison.

\begin{figure}[H]
    \centering
    \includegraphics[width=1\linewidth]{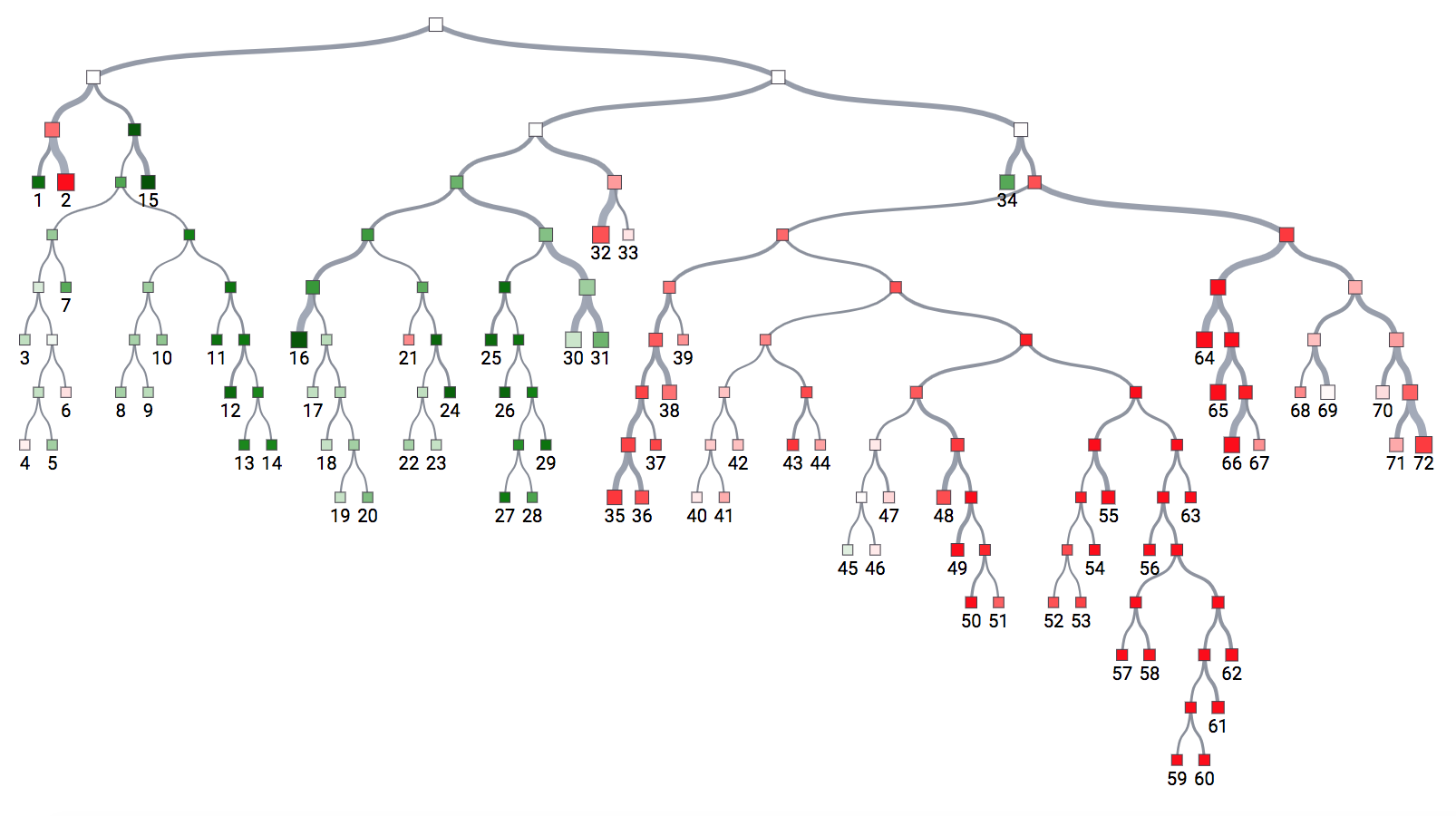}
    \caption{Discourse tree generated with our approach, containing 72 EDUs. Node color indicates inferred EDU-level sentiment.}
    \label{fig:teaser}
\end{figure}

In general it can be observed, that the MEGA-DT corpus dominates the comparison on the challenging but arguably useful inter-domain discourse parsing task. This suggests that the approach taken for the MEGA-DT treebank improves the structure prediction through more diversity in low-level trees, and properly addresses the nuclearity-prediction task.
To give further insight into the generation approach, we show a qualitative result of the ``silver-standard" annotation in Figure \ref{fig:teaser}.
The shown tree-structure indicates that generated trees are non-trivial, reasonably balanced and  strongly linked to the EDU-level sentiment.

In conclusion, our distant supervision approach enables the NLP community to extend any existing sentiment-annotated dataset with discourse-trees, allowing the automated creation of large-scale domain/genre-specific discourse treebanks.

\bibliography{emnlp2020}
\bibliographystyle{acl_natbib}

\end{document}